%% file: formatting-instructions-latex-2020.tex
\def\year{2021}\relax
\documentclass[letterpaper]{article} % DO NOT CHANGE THIS
\usepackage{aaai21}  % DO NOT CHANGE THIS
\usepackage{times}  % DO NOT CHANGE THIS
\usepackage{helvet} % DO NOT CHANGE THIS
\usepackage{courier}  % DO NOT CHANGE THIS
\usepackage[hyphens]{url}  % DO NOT CHANGE THIS
\usepackage{graphicx} % DO NOT CHANGE THIS
\usepackage{graphicx}  % DO NOT CHANGE THIS
\usepackage{mathtools}
 
\usepackage{dsfont}
\usepackage{enumitem}
\usepackage{caption}
\usepackage{multirow}
\usepackage{amsmath,amssymb} 

\usepackage{amsthm}
\theoremstyle{plain}% default

\newtheorem{theorem}{Theorem}

\usepackage[section]{placeins}

\usepackage{epsfig}
\usepackage{amsmath,amssymb} 
\usepackage{multirow}

\usepackage{algorithm}
\usepackage{algorithmic}

\title{Deep Verifier Networks: Verification of Deep Discriminative Models\\ with Deep Generative Models}

\author{Tong Che{$^{1\dag}$}, Xiaofeng Liu{$^{2\dag*}$}, Site Li{$^{3}$}, Yubin Ge{$^{4}$}, Ruixiang Zhang{$^{1}$},\\
{\Large\textbf{Caiming Xiong{$^{5}$}, Yoshua Bengio{$^{1}$} }}\\~\\
{$^{1}$}Mila, Universit de Montral~~~{$^{2}$}Harvard Medical School, Harvard University and Fanhan Tech.\\{$^{3}$}Carnegie Mellon University~~~ {$^{4}$}University of Illinois at Urbana-Champaign~~~{$^{5}$}Salesforce Research \\
{\small{{$^{\dag}$}Contribute equally~~{$^{*}$}Corresponding to: liuxiaofengcmu@gmail.com}}
}

\begin{document}

\maketitle

\input{0_Abstract.tex}

\input{1_Introduction.tex}

\input{2_RelatedWork.tex}

\input{3_Approach.tex}

\input{4_Experiments.tex}

\input{5_Conclusion.tex}

\bibliographystyle{aaai} \bibliography{egbib}

\end{document}

%% file: 0_Abstract.tex
\begin{abstract}

AI Safety is a major concern in many deep learning applications such as autonomous driving. Given a trained deep learning model, an important natural problem is how to reliably verify the model's prediction. In this paper, we propose a novel framework --- deep verifier networks (DVN) to detect unreliable inputs or predictions of deep discriminative models, using separately trained deep generative models. Our proposed model is based on 
the concise conditional variational auto-encoders with disentanglement constraints to separate the label information from the latent representation. We give both intuitive and theoretical justifications for the model. Our verifier network is trained independently with the prediction model, which eliminates the need of retraining the verifier network for a new model. We test the verifier network on both out-of-distribution detection and adversarial example detection problems, as well as anomaly detection problems in structured prediction tasks such as image caption generation. We achieve state-of-the-art results in all of these problems. 
\end{abstract}

%% file: 1_Introduction.tex
\section{Introduction}

Deep learning models provide state-of-the-art performance in various applications such as image classification \cite{krizhevsky2012imagenet,Wang2020automated}, caption generation \cite{xu2015show}, sequence modeling \cite{chung2014empirical,liu2018dependency} and machine translation \cite{xu2015show}. However, such performance is based on the assumption that the training and testing data are sampled from the same distribution \cite{goodfellow2016deep}. Without this assumption, deep learning models can fail silently by producing high confidence incorrect predictions even on completely unrecognizable or irrelevant inputs \cite{amodei2016concrete}. For instance, the models trained on MNIST can produce 91\% confidence on random noise images \cite{Hendrycks2016A}. Generally speaking, the behavior of a trained deep learning model on a slightly different test distribution is unpredictable. One such problematic case is also shown in Fig. {1}. Unfortunately, there is very little control over the test distribution in real-world deployments due to dynamically changing environments or malicious attacks \cite{guo2017calibration}. In fact, well calibrating the predictive uncertainty of DNNs is important for many authentication, medical and self-driving systems \cite{liu2019permutation,liu2017line,liu2019dependency,liu2020severity,liu2020importance,liu2020wasserstein,liu2019unimodal,liu2020unimodal,liu2020symmetric,liu2019conservative,liu2018ordinal,liu2018joint,Han_2020_CVPR_Workshops}.

\begin{figure}[t]
\centering
  \includegraphics[height=3.8cm]{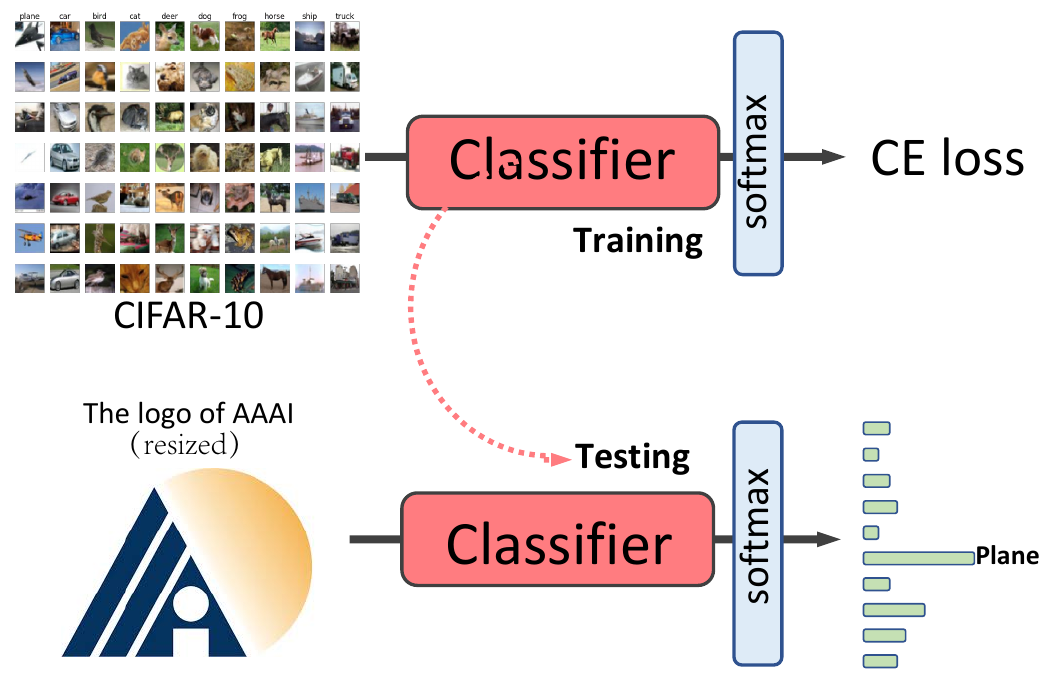} 
\caption{A network trained on CIFAR-10 will predict the resized 32$\times$32$\times$3 AAAI logo (OOD sample w.r.t. CIFAR-10) as the plane with high confidence.}\label{fig:arch} 
\end{figure}

Being overconfident on out-of-distribution (OOD) inputs has raised concerns about the safety of artificial intelligence (AI) systems. Recent research efforts try to address these concerns by developing models that can identify anomalous inputs, $i.e.,$ OOD samples \cite{amodei2016concrete}. Formally, the OOD detection problem can be formulated as a binary classification problem where the objective is to decide whether a test sample is from the training distribution ($i.e.,$ in-distribution, ID) or from a different distribution ($i.e.,$ OOD).  

\begin{table*}[t]\centering
\resizebox{\linewidth}{!}{
\centering
  \begin{tabular}{c|cccccccccccccc}
\hline
      &\cite{Hendrycks2016A}&\cite{Liang2018Enhancing}&\cite{Devries2018Learning} &\cite{Vyas2018Out}&\cite{Lee2018A} &\cite{choi2018waic}&\cite{Hendrycks2019deep} \\\hline
   1 & $\surd$ & -& -& -& -& -& -\\
   2 & $\surd$ & -& -& -& $\surd$&  -& -\\
   3 & -& -& -& -  & $\surd$& -& -\\
   4 & -& -& -& -& -&  -& -\\
\hline
  \end{tabular}
  }
  \caption{Summary comparison of the characteristics of the recent related methods.}
\end{table*}

In this paper, we propose to verify the predictions of deep discriminative models by using deep generative models that try to generate the input conditioned on the label selected by the discriminative model. We call this concept "deep verifier". The high-level idea is simple:  we train an inverse verification model $p(x|y)$ on the training data pairs $(x,y)$. Intuitively speaking, for an input-output pair $(x,y)$ with $y$ picked by the predictive model, we verify whether the input $x$ is consistent with $y$, by estimating if $p(x|y)$ is larger than a threshold. We design a density estimator of $p(x|y)$ using modified conditional VAEs. To ensure that the class code $y$ is not ignored as a conditioning variable, we impose a disentanglement constraint based on minimizing mutual information between latent variable representation $z$ and the label $y$. Although many different kinds of density estimators can be used in theory, we argue that the design of our model is robust to OOD samples and adversarial attacks, due to the use of latent variables with explicit and accurate density estimation.

Compared with previous approaches of OOD detection, our proposed method has 4 main advantages:

\noindent $\bullet$ 1.The verifier is trained independently of OOD distributions. Users do not need to figure out OOD samples before deployment of the system.

\noindent $\bullet$ 2.The verifier only needs to be trained once. No need to retrain the verifier for a new classifier. 

\noindent $\bullet$ 3.The verifier can detect ordinary OOD samples and malicious adversarial attacks in a unified manner.

\noindent $\bullet$ 4.The framework is very general, so that it applies to structured prediction problems as well, such as image captioning.

We summarize the comparison of these four advantages with previous methods in Table 1.

The proposed solution achieves the state-of-the-art performance for detecting either OOD or adversarial samples in all tested classification scenarios, and can be generalized well for structured prediction tasks ($e.g.,$ image caption). In Sec 3.4, we analysed why DVN is useful for both OOD and Adversarial examples.

%% file: 2_RelatedWork.tex
\section{Related Work}

Detecting the OOD samples in a low-dimensional space using traditional non-parametric density estimation, nearest neighbor and clustering analysis have been well-studied \cite{Pimentel2014A}. However, they are usually unreliable in high-dimensional spaces, e.g., images \cite{Liang2018Enhancing}.

OOD detection with deep neural networks has recently been an active research topic. \cite{Hendrycks2016A} found that trained DNNs usually have higher maximum softmax output for in-distribution examples than anomalous one. A possible improvement of this baseline is to consider both the in-distribution and out-of-distribution training samples during training \cite{Hendrycks2019deep}. However, enumerating all possible OOD distributions before deployment is usually not possible.

\cite{Liang2018Enhancing} proposed that the difference between maximum probabilities in softmax distributions on ID/OOD samples can be made more significant by using adversarial perturbation pre-processing during training. \cite{Devries2018Learning} augmented the classifier with a confidence estimation branch, and adjusted the objective using the predicted confidence score for training. \cite{Lee2017Training} trained a classifier simultaneously with a GAN, with an additional objective to encourage low confidence on generated samples. \cite{Hendrycks2019deep} proposed to use real OOD samples instead of generated ones to train the detector. \cite{Vyas2018Out} labels a part of training data as OOD samples to train the classifier, and they dynamically change the partition of ID and OOD samples. These improvements based on \cite{Hendrycks2016A} typically needs re-train a classifier with modified structures or optimization objectives. This can make it hard to maintain the original accuracy and is computationally expensive. 

\cite{Lee2018A} propose to obtain the class conditional Gaussian distribution, and then define confidence score using the Mahalanobis distance between the sample and the closest class-conditional Gaussian distribution. However, it also needs the input pre-processing and model change. Besides, many previous methods \cite{Liang2018Enhancing,Vyas2018Out,Lee2018A} need OOD samples for hyper-parameter ($e.g.,$ threshold for verification) selection, and these are usually not accessible. 

The main difference of our model with Bayesian NN based calibration models \cite{nalisnick2019detecting,snoek2019can,yao2019quality,guo2017calibration} is that our model does not need to modify the training procedure of the classifier. Bayesian NNs are notoriously hard and computationally expensive to train, and they need to be carefully designed and the model itself needs to be modified, which seriously limits their applications to real-world problems.

Recently, \cite{choi2018waic} proposed an unsupervised OOD detector by estimating the Watanabe-Akaike Information Criterion. The goal of our model is different from WAIC in that rather than just detecting OOD samples, DVNs aim to verify the predictions of a supervised predictive model, i.e., estimating $p(x|y)$ not just $p(x)$. We argue that modeling $p(x|y)$ is usually easier than directly modeling $p(x)$ as the former distribution contains less modes. 

Another motivation for modelling $p(x|y)$ instead of $p(x)$ is that for an adversarial attack and its classifier prediction $(x',y')$, it is usually much easier to verify $x'$ is not in $p(x|y')$ than to verify $x'$ is not in $p(x)$. For an adversarial attack $(x',y')$ modified from $(x,y)$, $y'$ suppose to be a different class from $y$. However, $x'$ is visually very much like $x$, namely $|x'-x|_{L^1}<\epsilon$. Therefore, given a wrong class $y'$, $p(x'|y')$ can be very easily verified to be small, since $x'$ is very close to a image in class $y$ while it should be very different from an image in class $y'$.

%% file: 3_Approach.tex
 \section{Methodology}

This paper targets the problem of verification of deep predictive models, as follows. Let $x\in\mathcal{X}$ be an input and $y\in\mathcal{Y}$ be the ground-truth value to be predicted. The in-distribution examples are sampled from the joint data-generating distribution $p_{\text{in}}(x,y)=p_{\text{in}}(y|x)p_{\text{in}}(x)$. We propose to reverse the order of the prediction process of $p(y|x)$ and try to compute the conditional probability $p(x|y)$, where $y$ is the label value guessed by the classifier to be verified (e.g., the one
with the highest probability according to the deep network). We evaluate whether the input $x$ is consistent with that $y$.

The predictive model to be verified $p_{\theta}(y|x)$ is trained on a dataset drawn from the $p_{\text{in}}(x,y)$, and may encounter samples from both $p_{\text{in}}(x,y)$ and $p_{\text{out}}(x,y)$ ($i.e.,$ out-of-distribution or adversarial samples) at test time. Note there is some subtle difference between OOD (unlikely under $p_{\text{in}}(x)$) and adversarial examples (unlikely under the ground truth joint, but with high $p_{\text{in}}(x)$, especially if a small amount of noise is allowed).

Our goal is to verify if the pair $(x,y)$ for $y$ guessed by the predictive model given $x$ is consistent with $p_{\text{in}}(x,y)$. We train a verifier network $q_\phi(x|y)$ as an approximation to the inverse posterior distribution $p(x|y)$. Modelling $p(x|y)$ instead of $p(x)$ as a verification has many advantages: (1) Usually $p(x)$ is much more diverse than the conditional distribution $p(x|y)$, so modelling $p(x|y)$ is much easier than modelling $p(x)$. (2) Modelling $p(x|y)$ allows us to provide a unified framework for verifying OODs, adversarial examples, and mis-classifications of the classifier. 

\subsection{Basic Model}

\begin{figure}[t]
\centering
  \includegraphics[height=4cm]{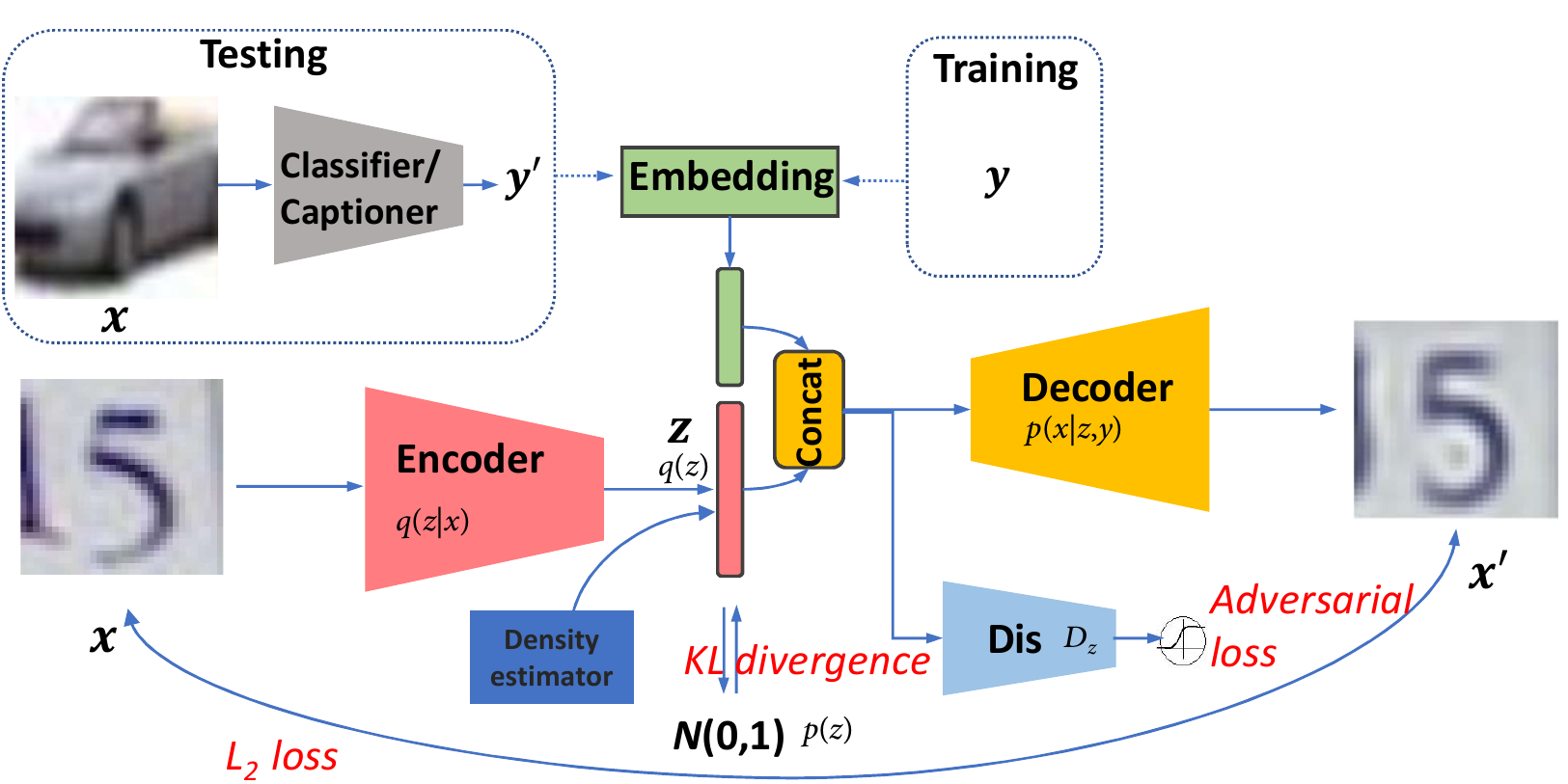}
\caption{The architecture of our Deep Verifier Network (DVN). We use ground-truth label $y$ of training example $x$ in training while using the trained model prediction $y'$ of testing image in testing.}\label{fig:arch}
\end{figure}

\begin{table*}[t]
\centering
\caption{OOD verification results of image classification under different validation setups. All metrics are percentages and the best results are bolded. The backbone classifier in SUF and our DVN is ResNet34 ({He et al. 2016}), while ODIN uses more powerful wide ResNet40 with width 4 ({Zagoruyko et al. 2016}).}
\resizebox{\linewidth}{!}{
\centering
  \begin{tabular}{cc|ccc|ccc}
\hline
  &&\multicolumn{3}{c}{Validation on OOD samples}&\multicolumn{3}{|c}{Validation on adversarial samples}\\\cline{3-8}
In-Dist&OOD& \multicolumn{1}{c|}{TNR@TPR 95\%}  &\multicolumn{1}{c|}{AUROC} & {Verification acc.}&\multicolumn{1}{c|}{TNR@TPR 95\% } &\multicolumn{1}{c|}{AUROC} &Verification acc.\\\cline{3-8}

&&\multicolumn{3}{c}{ ODIN / SUF / \textbf{Our DVN} / Glow based DVN / Pixel CNN based DVN}&\multicolumn{3}{|c}{ ODIN / SUF /\textbf{Our DVN} / Glow based DVN / Pixel CNN based DVN}\\\hline

CIFAR-10&SVHN&86.2/90.8/\textbf{96.4}/95.1/94.7&95.5/98.1/\textbf{99.0}/98.2/98.0 &91.4/93.9/\textbf{95.1}/93.7/93.9 &70.5/89.6/\textbf{95.2}/93.8/91.0 &92.8/97.6/\textbf{98.1}/97.5/97.6 &86.5/92.6/\textbf{94.2}/93.1/93.4\\
DenseNet&T-ImageN&92.4/95.0/\textbf{96.2}/95.1/94.8 &98.5/98.8/\textbf{99.0}/98.4/98.2 &93.9/95.0/\textbf{97.3}/96.4/96.6 &87.1/94.9/\textbf{95.6}/94.7/94.3 &97.2/98.8/\textbf{99.1}/98.8/98.6 &92.1/95.0/\textbf{97.4}/96.5/96.2 \\
&LSUN&96.2/97.2/\textbf{98.6}/97.5/97.3&99.2/\textbf{99.3}/\textbf{99.3}/98.9/98.9&95.7/96.3/\textbf{96.8}/96.2/96.0&92.9/97.2/\textbf{97.9}/97.2/97.3&98.5/99.2/\textbf{99.3}/98.7/98.8&94.3/96.2/\textbf{97.5}/96.6/96.3\\\hline

CIFAR-100&SVHN&70.6/82.5/\textbf{95.2}/93.0/92.8&93.8/97.2/\textbf{97.3}/97.1/96.8&86.6/91.5/\textbf{93.4}/92.4/92.5&39.8/62.2/\textbf{90.5}/85.7/86.0&88.2/91.8/\textbf{92.2}/90.9/91.0&80.7/84.6/\textbf{86.3}/85.4/85.7\\
DenseNet&T-ImageN&42.6/86.6/\textbf{99.0}/96.4/96.5&85.2/{97.4}/\textbf{99.4}/96.8/95.6&77.0/92.2/\textbf{98.8}/95.8/95.0&43.2/87.2/\textbf{99.1}/98.5/98.5&85.3/97.0/\textbf{97.8}/96.9/96.4&77.2/91.8/\textbf{98.0}/96.3/95.0\\
&LSUN&41.2/91.4/\textbf{93.7}/92.5/93.1&85.5/98.0/\textbf{98.2}/97.6/97.5&77.1/93.9/\textbf{99.9}/98.0/98.2&42.1/91.4/\textbf{98.6}/97.8/96.0&85.7/97.9/\textbf{98.3}/97.9/97.8&77.3/93.8/\textbf{95.4}/94.2/94.6\\\hline

SVHN&CIFAR-10&71.7/96.8/\textbf{97.4}/95.7/96.2&91.4/98.9/\textbf{99.2}/98.8/98.2&85.8/95.9/\textbf{96.5}/95.1/95.0&69.3/97.5/\textbf{97.8}/97.4/97.0&91.9/98.8/\textbf{99.1}/98.1/98.0~&86.6/96.3/\textbf{97.4}/96.6/96.7\\

DenseNet&T-ImageN&84.1/99.9/\textbf{100}/98.3/98.0&95.1/\textbf{99.9}/\textbf{99.9}/98.5/98.4&90.4/98.9/\textbf{99.2}/98.0/97.7&79.8/\textbf{99.9}/\textbf{99.9}/96.4/98.3&94.8/99.8/\textbf{99.9}/96.7/97.1&90.2/98.9/\textbf{99.4}/97.6/97.7\\
&LSUN&81.1/\textbf{100}/\textbf{100}/98.7/98.5&94.5/\textbf{99.9}/\textbf{99.9}/97.9/98.2&89.2/99.3/\textbf{99.6}/98.8/98.4&77.1/\textbf{100}/\textbf{100}/98.2/98.5&94.1/99.9/\textbf{100}/96.8/96.5&89.1/99.2/\textbf{99.5}/97.2/98.1\\\hline

CIFAR-10&SVHN&86.6/96.4/\textbf{98.4}/97.3/97.0&96.7/{99.1}/\textbf{99.2}/98.5/98.6&91.1/95.8/\textbf{97.3}/96.2/96.1&40.3/75.8/\textbf{98.5}/97.6/97.4&86.5/95.5/\textbf{96.1}/95.5/95.3&77.8/89.1/\textbf{92.2}/91.2/91.0\\
ResNet&T-ImageN&72.5/97.1/\textbf{98.0}/97.0/96.9&94.0/99.5/\textbf{99.6}/98.5/98.5&86.5/96.3/\textbf{96.9}/94.7/94.9&96.6/95.5/\textbf{97.1}/96.2/95.9&93.9/99.0/\textbf{99.2}/98.3/98.1&86.0/95.4/\textbf{96.3}/95.5/95.1\\
&LSUN&73.8/98.9/\textbf{99.0}/97.6/97.7&94.1/\textbf{99.7}/\textbf{99.7}/97.8/97.5&86.7/97.7/\textbf{97.9}/96.3/96.0&70.0/98.1/\textbf{98.9}/96.8/96.5&93.7/\textbf{99.5}/\textbf{99.5}/97.6/97.7&85.8/97.2/\textbf{98.0}/96.8/96.5\\\hline

CIFAR-100&SVHN&62.7/91.9/\textbf{98.5}/96.5/96.6&93.9/98.4/\textbf{98.8}/98.3/98.0&88.0/93.7/\textbf{94.8}/92.9/93.2&12.2/41.9/\textbf{86.2}/82.4/83.5&72.0/84.4/\textbf{86.3}/84.7/84.2&67.7/76.5/\textbf{89.4}/87.5/87.4\\
ResNet&T-ImageN&49.2/90.9/\textbf{97.2}/95.6/95.3&87.6/98.2/\textbf{98.5}/98.0/97.7&80.1/93.3/\textbf{94.3}/93.0/93.1&33.5/70.3/\textbf{94.6}/92.2/91.8&83.6/87.9/\textbf{90.3}/86.6/86.5&75.9/84.6/\textbf{89.8}/85.2/85.8\\
&LSUN&45.6/90.9/\textbf{99.3}/98.5/98.8&85.6/98.2/\textbf{98.6}/96.7/97.0&78.3/93.5/\textbf{98.7}/96.9/95.7&31.6/56.6/\textbf{93.5}/90.2/90.1&81.9/82.3/\textbf{95.2}/92.9/92.8&74.6/79.7/\textbf{91.9}/90.0/88.9\\\hline

SVHN&CIFAR-10&79.8/98.4/\textbf{99.4}/97.9/97.5&92.1/99.3/\textbf{99.9}/98.1/98.2&89.4/96.9/\textbf{97.5}/96.3/96.3&79.8/94.1/\textbf{94.5}/93.7/93.5&92.1/97.6/\textbf{98.7}/96.5/96.2&89.4/94.6/\textbf{94.8}/93.7/93.0\\
ResNet&T-ImageN&82.1/99.9/\textbf{100}/98.5/98.4&92.0/\textbf{99.9}/\textbf{99.9}/96.3/96.5&89.4/99.1/\textbf{99.2}/95.8/96.7&80.5/99.2/\textbf{99.7}/98.5/98.3&92.9/99.3/\textbf{99.5}/97.2/97.0&90.1/98.8/\textbf{99.3}/98.0/98.2\\
&LSUN&77.3/\textbf{99.9}/\textbf{99.9}/96.4/96.4&89.4/\textbf{99.9}/\textbf{99.9}/97.6/97.4&87.2/99.5/\textbf{100}/99.0/98.9&76.3/\textbf{99.9}/\textbf{99.9}/96.5/97.4&90.7/\textbf{99.9}/99.8/96.8/96.7&88.2/99.5/\textbf{99.8}/98.3/98.1\\
\hline
  \end{tabular}
}
\end{table*}

Our basic model is a conditional variational auto-encoder shown in Fig. {2}. The model is composed of two deep neural networks, a stochastic encoder $q(z|x)$ which takes input $x$ to predict a latent variable $z$ and a decoder $p(x|z,y)$ which takes both latent variable $z$ and the label $y$ to reconstruct $x$. 
One problem with training of conditional variational auto-encoders is that the decoder can ignore the effect of input label $y$, passing all information through the continuous latent variable $z$. This is not desirable as we want to use the decoder to model the conditional likelihood $p(x|y)$, not $p(x)$. Hence in this paper, we train the encoder so that it outputs a $z$ which is approximately independent of $y$.
The encoder and decoder are thus jointly trained to maximize the evidence lower bound (ELBO): \begin{equation}
  \log p(x|y) \geq {\mathbb{E}}_{q(z|x)}[\log p(x|z,y)]-\text{KL}(q(z|x)||p(z)) \label{eq:1}
\end{equation}
The equality holds iff $q(z|x)= p(z|x,y)$, where $p(z|x,y)$ is the ground truth posterior. We note that the conditional GAN is not applicable here since its objective does not optimize the likelihood \cite{liu2019feature,liu2020auto3d,liu2020disentanglement,He_2020_CVPR_Workshopsa,He_2020_CVPR_Workshopsb}. 

\subsection{Disentanglement constraints for anomaly detection}
To achieve this independence, we propose to add a disentanglement penalty to minimize the mutual information between $z$ and $y$.  Namely, besides the ELBO loss, we also minimize the mutual information estimator $\hat{I}(z,y)$ together with the loss, yielding: \begin{equation}
  L =  -{\mathbb{E}}_{q(z|x)}[{\log} p(x|z,y)+\lambda \hat{I}(y,z)]+\text{KL}(q(z|x)||p(z)) \label{eq:3}
\end{equation}

In this paper, we use deep Infomax \cite{hjelm2018learning} as the proxy for minimizing the mutual information (MI) between $z$ and $y$. The mutual information estimator is defined as: \begin{equation}
   \hat{I}(z,y) = \mathbb{E}_{p(y,z)}[-s_+(-T(y,z))]-\mathbb{E}_{p(y)p(z)}[s_+(T(z,y))]
\end{equation}
where $s_+$ is the softplus function and $T(y,z)$ is a discriminator network. Just like GAN discriminators, $T$ is trained to maximize $\hat{I}(y,z)$, in order to get a better lower-bound estimation of the (JS-version) mutual information, while $L$ (and in particular the encoder
and decoder) is optimized (considering $T$ fixed) to minimize $\hat{I}(y,z)$.

\subsection{Measuring the likelihood as anomaly score}

Our anomaly verification criterion is based on estimating the log-likelihood $\log p(x|y)$ for test samples. Importance sampling is a possible solution to provide an unbiased estimate of $p(x|y)$ when we have a VAE. Following IWAE \cite{burda2015importance}, the $k$-sample importance weighting estimate of the log-likelihood is a lower bound of the ground truth likelihood $\mathcal{L}(x|y)=\mathbb{E}_{x\sim p(\cdot |y)}[\log p(x|y)]$: \begin{equation}
\mathcal{L}_k(x|y)=\mathbb{E}_{z_1,...,z_k \sim q(z|x)}  [ \log \frac{1}{k} \sum_{i=1}^{k} \frac{p(x,z_i|y)}{q(z_i|x)}].
\label{eq:Lk}
\end{equation}
where $q(z)$ is a corrected density described below.
We use the fact that $\mathcal{L}_k(x|y)\rightarrow \mathcal{L}(x|y)$ as $k\rightarrow \infty$ to estimate the likelihood. As will be discussed below, we want the decoder $p(x|z,y)$ be evaluated on the same input distribution for $z$ as it is trained, which is not exactly
the original Gaussian prior $p(z)$, so we will form a refined estimator of the prior, denoted $p^*(z)$. The quantities $\mathcal{L}_k(x|y)$ form a monotonic series of lower bounds of the exact log-likelihood $\log p(x|y)$,
with $\mathcal{L}_1\leq \mathcal{L}_2\leq \cdots \mathcal{L}_k\leq \log p(x|y)$). They have the property that when $k\rightarrow \infty$, $\mathcal{L}_k\rightarrow \log p(x|y)$. In our experiments we chose $k=100$ for a good approximation of the exact likelihood.

In our algorithm, the distribution of $z$ actually fed into decoder $p(x|z,y)$ during training is $q(z)=\int q(z|x)p_d(x)dx$. However, this distribution $q(z)$ can be drastically different from the Gaussian prior $p(z)$. So instead of using the Gaussian $p(z)$ as a prior for the decoder network in Eq.~\ref{eq:Lk}, we use $q(z)$ and estimate the corrected likelihood of $x$ under this directed generative model, as $p(x,z|y)= q(z)p(x|z,y)$. In order to estimate the density of $q(z)$, we propose to train an additional discriminator $D_z$ to distinguish $p(z)$ and $q(z)$. ${D_z}$ is trained to discriminate the real distribution of latent variable $q(z)=\int p_d(x)e(z|x)dx$ ($p_d(x)$ is the data distribution of $x$, $e(z|x)$ is the encoder network) and Gaussian prior distribution $p(z)$, with ordinary GAN loss \cite{goodfellow2014generative,liu2018normalized,liu2017adaptive,liu2018data,liu2019hard}. Both $q(z)$ and $p(z)$ are easy to sample, so a discriminator is easy to train with the samples. In the GAN, the optimal discriminator $D_z$ can be $D_z=\frac{p(z)}{p(z)+q(z)}$ \cite{goodfellow2016nips}. After $D_z$ is trained (in theory optimally) and since $p(z)$ is known ($i.e.,$ Gaussian), we can estimate $q(z)=\frac{1-D_z(z)}{D_z(z)}p(z)$.

We classify a sample $x$ as an OOD sample if the log-likelihood is below the threshold $\delta$ and the $x$ is an in-distribution sample, otherwise. \begin{equation}
x\in\left\{
        ~  \begin{array}{lr}
             \text{in-distribution (ID)},~~~~~~~{\rm if}\;\; L_k  \geq \delta&  \\
             \text{out-of-distribution (OOD),~~~otherwise}&  
             \end{array}
\right.
\end{equation}
We set $\delta$ to the threshold corresponding to 95\% true positive rate (TPR), where the TPR refer to the probability of in-distribution validation samples are correctly verified as the in-distribution. Therefore, the threshold selection in our model is only tuned on in-distribution validation datasets. This differentiates our method with the other threshold based detector which need the OOD samples for hyper-parameter validation \cite{Liang2018Enhancing,Lee2018A}. We note that the distribution of OOD samples is usually not accessible before the system deployment.

\subsection{Theoretical Justification}
The loss function we optimize can be written as: 
\begin{equation}
\begin{aligned}
    L = &L_1 + \lambda L_2 = \mathbb{E}_{x,y\sim p_d}[-\mathbb{E}_{q(z|x)}[{\log} p(x|z,y)] \\&+\text{KL}(q(z|x)||p(z))+\lambda \mathbb{E}_{q(z|x)}[\hat{I}(y,z)]] \label{eq:4}
\end{aligned}\end{equation} where $p(x|z,y)$ is the decoder we are training. In this section, we use the following convention. Symbol $p$ means probability distributions induced by the decoder, and symbol $q$ means probability distributions induced by the encoder. Also denote $p_d$ for real data distributions. Specifically, we define joint distribution $q(z,x,y)=q(z|x)p_d(x,y)$\footnote{In this paper we assume $q(z|x)=q(z|x,y)$, the motivation is during test time, $y$ may be a wrong label, we don't want it to confuse the encoder. See detailed ablation in our Appendix.}. We have the following theorem that justifies the two parts of the above loss.

\begin{theorem}
 
(i) $-L_1$ is a variational lower bound of $\mathbb{E}_{x,y\sim p_d}[\log p(x|y)]$.The bound is tight when $q$ is expressive enough and $z,y$ are conditionally independent given $x$. 

\noindent(ii) If we have $I(y,z)=0$, where $(y,z)\sim \mathbb{E}_{x\sim p_d}$ $[p_d(y|x)q(z|x)]$ (namely $L_2\approx 0$), and assume that the decoder is perfect in sense that $p(x|y,z)=q(x|y,z)$, then we have our evaluation metric $\mathbb{E}_{z\sim q(z)}[p(x|y,z)]=p_d(x|y)$.
Namely, if $I(y,z)=0$, and the decoder is trained to optimal, , then no matter what the encoder looks like, the likelihood estimator  we are using is $\mathbb{E}_{z\sim q(z)}[p(x|y,z)]$ is equal to the groundtruth likelihood. 

\end{theorem}

This justifies why we need $L_2$ loss. Note that even with an encoder mapping everything to zero, the claim $\mathbb{E}_{z\sim q(z)}[\log p(x|y,z)]$ still equals to the ground truth likelihood. In this case, $\log p(x|y,z)=\log p(x|y)$ and is a constant with respect to $z$.

\subsection{Intuitive Justifications}\label{subsection1}
We now present an intuitive justification for the above algorithm. First, consider the following part of our training loss: \begin{equation}
  L_1 = -{\mathbb{E}}_{q(z|x)}[{\log} p(x|z,y)]+\text{KL}(q(z|x)||p(z)) \label{eq:2}
\end{equation}

It is well known that deep neural networks can generalize well for in-distribution samples, but their behavior out-of-distribution is less clear. Suppose $x$ is an out-of-distribution sample, with $y$ be the corresponding output of the classifier. Then the behavior of the stochastic encoder $q(z|x)$ is undefined. We denote $q(z)=\int q(z|x)p_d(x)$ the distribution to train $q(x|y,z)$. There are two cases: (1) $q(z|x)$ maps $x$ to $z$ with low density in $q(z)$. This case can be easily detected because $q(z)$ is easily computable. In this case the second term in Eq. \ref{eq:2} is a large negative number. (2) $q(z|x)$ maps $x$ to $z$ with high density in $q(z)$. Then since we train the decoder network with the input distribution $q(z)$ and because $y$ and $z$ are approximately independent, so $(z,y)$ looks like an in-distribution input for decoder $p(x|z,y)$. Thus $p(x|y,z)$ should map to some in-distribution $x'$ with class label $y$. Since input $x$ is an OOD sample and reconstruction $x'$ is an in-distribution sample, the reconstruction has to be bad. In this case, the first term in Eq. \ref{eq:2} is a large negative number. So in both cases, the log-likelihood score $L_k$ derived from our model should be a large negative number. This is why our model is robust to both adversarial and OOD samples. 

\subsection{Replacing VAEs with Other Density Estimators?}
In theory, we can use any other density estimator besides our modified conditional VAE (such as auto-regressive models and flow-based models) to estimate $p(x|y)$. However, our experiments and previous observations suggest that these other models may have drawbacks that would make them less suitable for this task. The comparison with the DVN that is based on PixelCNN \cite{van2016conditional} and Glow \cite{kingma2018glow} are compared in Tab. {2}, which is consistently inferior than our VAE solution. Auto-regressive models are quite slow and may ignore the conditioning label $y$ \cite{bowman2015generating}. Flow-based models were found to be less robust to adversarial examples, assigning higher likelihood on OOD samples than in-distribution samples \cite{nalisnick2018deep}. We have intuitively explained in last subsection about why our modified cVAE based model does not suffer from the same problem as flow-based models, thanks to our disentanglement regularizer, which relies on the existence of a latent space.

%% file: 4_Experiments.tex
\begin{figure}[t]
\centering
 \includegraphics[height=4.5cm]{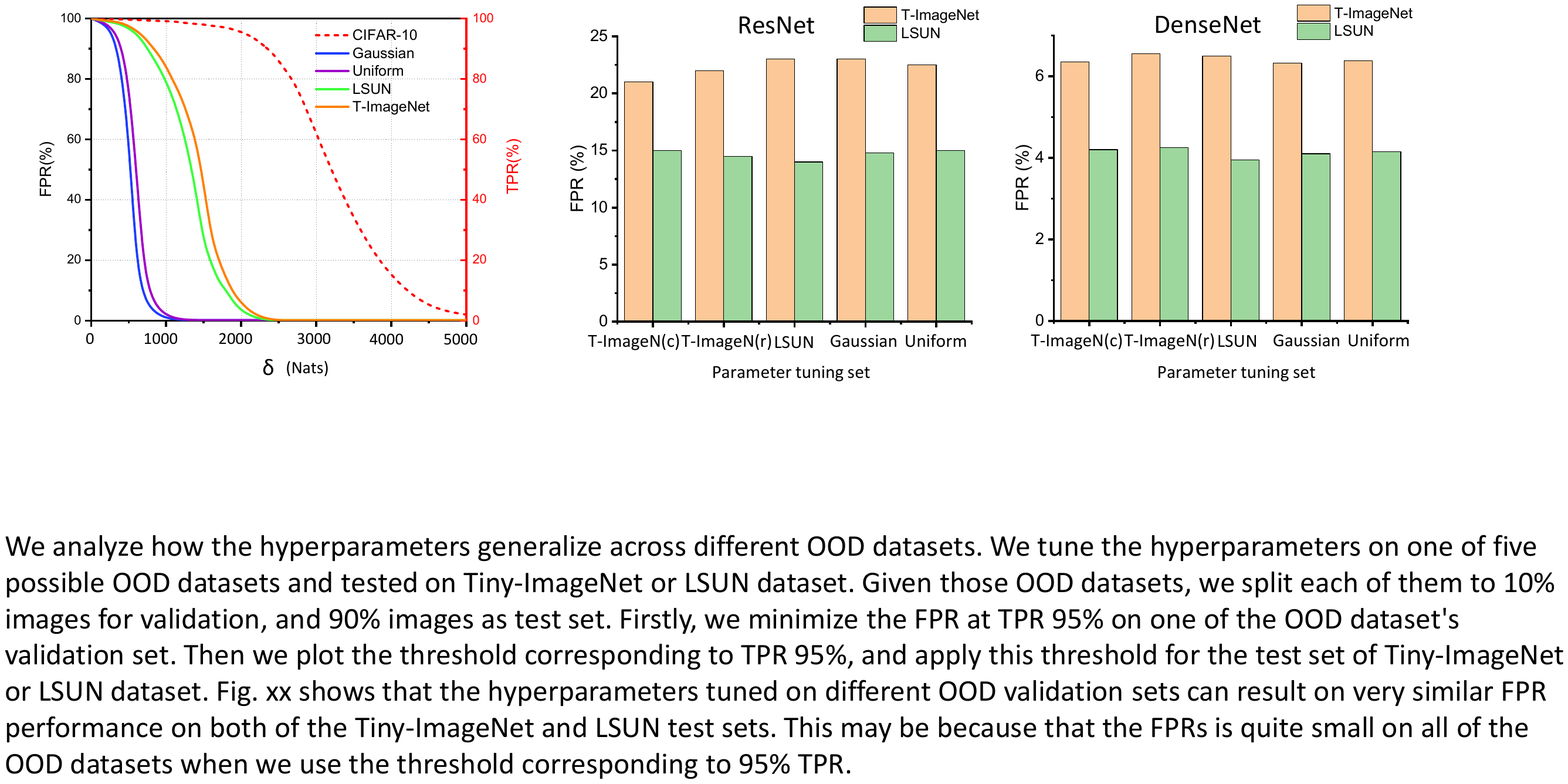}
  \caption{FPR (for OOD) and TPR (for ID) under different $\delta$ when using CIFAR-10 as the in-distribution dataset, and use Tiny-ImageNet(resize), LSUN and Gaussian/Uniform noise as OOD. CIFAR-10 only applicable to the TPR which use the dashed red line and indicated by the right axis while the other OOD datasets use the left FPR axis.}\label{figdelta}
\end{figure}

\section{Experimental results}

\begin{figure}[t]
\centering
\includegraphics[height=3.5cm]{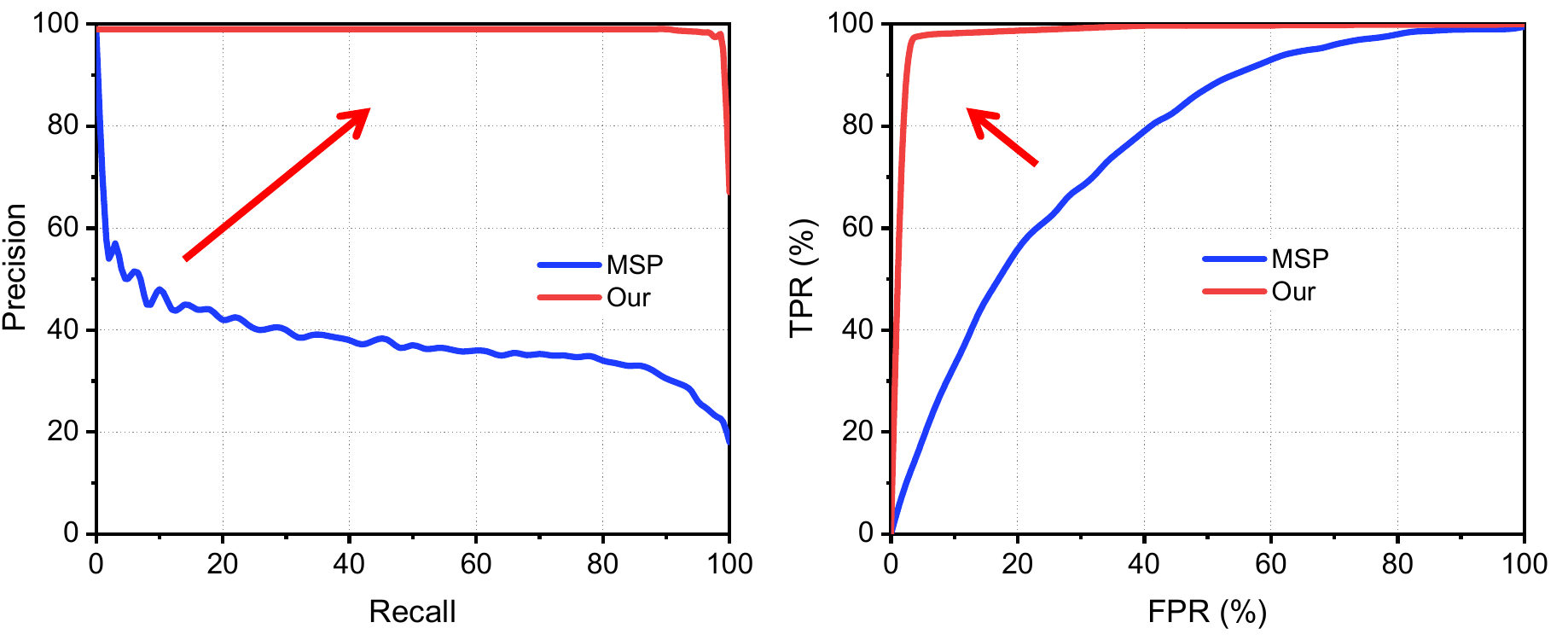} 
  \caption{Comparison with baseline MSP \cite{Hendrycks2016A} using DenseNet, with Tiny-ImageNet as in-distribution and LSUN as OOD.}\label{fige1} 
\end{figure}

In this section, we demonstrate the effectiveness of the proposed DVN on several classification benchmarks, and show its potential for the image captioning task. We choose the DenseNet ({Huang et al. 2017}) and ResNet ({He et al. 2016}) architectures as the backbones of our experiments. Following the definitions in previous work, we denote the in-distribution images as positive samples, while the OOD ones as the negative samples.

For evaluation, we measure the True Negative Rate or False Positive Rate at {95\%} True Positive Rate (i.e., {TNR@TPR95\%} or {FPR@TPR95\%}), Area under the receiver operating characteristic curve ({AUROC}), Area under the precision-recall curve ({AUPR}) and {Verification accuracy}. We detailed these metrics in Supplementary Materials. Noticing that AUROC, AUPR and verification accuracy are threshold ($\delta$)-independent evaluation metrics.

\begin{table*}[t]
\centering
\caption{Comparison of AUROC (\%) under different validation setups. The best results are bolded. We also compared the use of negative samples for training and input image pre-processing.}
\resizebox{0.8\linewidth}{!}{
  \begin{tabular}{c|c|c|c|c|c|c|c|c|c|c|c|c|c|c|c}
  \hline
  \multirow{2}{*}{~} & \multirow{2}{*}{Dataset} & \multirow{2}{*}{Method} & Negative& Pre-  & {Deep} & \multirow{2}{*}{CW} &  \multirow{2}{*}{BIM}&\multirow{2}{*}{~} & \multirow{2}{*}{Dataset} & \multirow{2}{*}{Method} & Negative& Pre-  & {Deep} & \multirow{2}{*}{CW} &  \multirow{2}{*}{BIM}\\ 
  &&& Sample & proce &Fool&&&&&& Sample & proce &Fool&\\\hline\hline

\multirow{12}{*}{\rotatebox{90}{DenseNet}}& &KD+PU & FGSM &-&68.34 & 53.21 & 3.10&\multirow{12}{*}{\rotatebox{90}{ResNet}}& &KD+PU & FGSM&- & 76.80 &56.30 & 16.16\\
& CIFAR & LID & FGSM &-& 70.86 & 71.50 & 94.55 && CIFAR & LID & FGSM&- &71.86&77.53&95.38  \\
&-10 &SUF & FGSM & Yes &87.95 & 83.42 & {99.51}&&-10& SUF & FGSM& Yes &78.06&93.90&98.91 \\
& &Our & - & - &\textbf{96.14} & \textbf{96.38} & \textbf{99.82}&&& Our & - & - &\textbf{95.45} & \textbf{99.51} & \textbf{99.57}\\\cline{2-8} \cline{10-16}

&& KD+PU & FGSM &-& 65.30 & 58.08 & 66.86& && KD+PU & FGSM&- & 57.78 & 73.72 & 68.85\\
& CIFAR & LID & FGSM&-  &69.68&72.36 & 68.62& & CIFAR & LID & FGSM&-     & 63.15& 75.03& 55.82  \\
&-100 &SUF & FGSM& Yes &75.63&86.20&98.27&&-100& SUF & FGSM& Yes &81.95& 90.96&96.38\\
& &Our & - & - &\textbf{97.01} & \textbf{98.55} & \textbf{99.94}&&& Our & - & - &\textbf{97.22} & \textbf{99.38} & \textbf{99.72}\\\cline{2-8}\cline{10-16}

&& KD+PU & FGSM&- & 84.38 &82.94 & 83.28&& &KD+PU & FGSM&- & 84.30 & 67.85 & 43.21\\
& SVHN & LID & FGSM&- &80.14&85.09&92.21 & & SVHN & LID & FGSM&-   &67.28&76.58&84.88    \\
& &SUF & FGSM& Yes &93.47&96.95&{99.12}&& &SUF & FGSM& Yes &72.20&86.73&95.39 \\
& &Our & - & - &\textbf{98.14} & \textbf{99.35} & \textbf{100.00}& & &Our & - & - &\textbf{97.13} & \textbf{99.76} &\textbf{100.00}\\
\hline
\end{tabular}}
 
\end{table*}

\subsection{Detecting OOD samples for classification}

~~~~\textbf{Datasets.} The Street View Housing Numbers (\textbf{SVHN}) dataset ({Netzer et al. 2011}) consists of color images depicting house numbers, which range from 0 to 9. Images have a resolution of 32$\times$32. For our tests, we use the official training set split which contains 73,257 images, and the test set split, which has 26,032 images. The \textbf{CIFAR-10/100} dataset ({Krizhevsky and Hinton. 2009}) consists of 10/100 classes colour images. The training set has 50,000 images, while the test set has 10,000 images. The dataset\footnote{https://tiny-imagenet.herokuapp.com/} is a subset of the ImageNet dataset  ({Deng et al. 2009}). Its test set contains 10,000 images from 200 different classes. It contains the original images, downsampled to 32$\times$32 pixels. The Large-scale Scene UNderstanding dataset (\textbf{LSUN}) ({Yu et al. 2015}) has a test set with 10,000 images from 10 different classes. The LSUN (crop) and LSUN (resize) are created in a similar downsampling manner to the TinyImageNet datasets. The \textbf{Uniform noise} and \textbf{Gaussian noise} dataset are with 10,000 samples respectively, which are generated by drawing each pixel in a 32$\times$32 RGB image from an i.i.d uniform distribution of the range [0, 1] or an i.i.d Gaussian distribution with a mean of 0.5 and variance of 1 \cite{Liang2018Enhancing}.

\textbf{Setups.} For fair comparisons, the backbones of the classifiers used here are the 100-layer DenseNet with growth rate 12 \cite{Liang2018Enhancing,Lee2018A} and 34-layer ResNet \cite{Lee2018A}. They are trained to classify the SVHN, CIFAR-10, CIFAR-100 and Tiny-ImageNet datasets, of which test set is regarded as the in-distribution dataset in our testing stage. The dataset different from its training dataset is considered as OOD. We use four convolution or deconvolution layers for the encoder and decoder structure, and $z$ is a 128-dimension vector. The discriminator is a two-layer fully connected layer network with sigmoid output and binary cross-entropy loss. The hyper-parameters in previous methods \cite{Liang2018Enhancing,Lee2018A} need to be tuned on a validation set with 1,000 images from each in-distribution and OOD pair. We note that the threshold of the DVN is tuned on in-distribution only. This corresponds to a more realistic scenario, since the OOD nature of real-world applications is usually uncontrollable.

\textbf{{Effects of the threshold and performance across datasets}.} How the hyper-parameters ($e.g.,$ $\delta$) generalize across different OOD datasets is a challenging aspect of the system deployment. Most of the previous methods require a small set of OOD samples, with $\delta$ calibrated by evaluating the verification error at different values of $\delta$. However, the more realistic scenario is that we do not have access to the OOD examples in the testing stage. A promising trend is improving the performance on an unknown OOD when using the model tuned on a similar OOD \cite{Liang2018Enhancing,Lee2018A}. We argue that our DVN is essentially free from such worries, since it does not need any OOD sample in the validation. To investigate how the threshold affects the FPR and TPR, Fig. \ref{figdelta} shows their relationship when training on CIFAR-10 and different OOD datasets are used in the test stage, with a DenseNet backbone. Note that the TPR (red axis) is used for in-distribution dataset CIFAR-10 (red dashed line), while FPR is used for OODs. We can observe that the threshold corresponding to 95\% TPR can produce small FPRs on all OOD datasets. When the OOD images are sampled from some simple distributions ($e.g.,$ Gaussian or Uniform), the available window of threshold $\delta$ can be larger.

\begin{table}[t]
\centering
\caption{Test error rate of classification on CIFAR-10/100 using DenseNet as backbone. Our DVN does not re-train or modify the structure of the original trained classifier.}
\resizebox{0.7\columnwidth}{!}{
 
	\begin{tabular}{c|cc}
\hline
&CIFAR-10&CIFAR-100\\\hline\hline
ODIN/SUF&4.81&22.37\\\hline
DenseNet/DVN&\textbf{4.51}&\textbf{22.27}\\
\hline
	\end{tabular}}  \label{table1} 
\end{table}

\textbf{Comparison with SOTA.} The main results are summarised in Tab. {2}. For each in\&out-of-distribution pair, we report the performance of ODIN \cite{Liang2018Enhancing}, SUF \cite{Lee2018A} and our DVN. Notably, DVN consistently outperforms the previous methods and achieves a new state-of-the-art. As shown in Tab. {3}, the pre-processing and model change in ODIN and SUF increase the error rate of the original classifier for the in-distribution test, while DVN does not affect the classification accuracy on the accepted in-distribution datasets (i.e., CIFAR-10 or 100), when the OOD examples are not presented.

Considering the technical route of DVN is essentially different from ODIN and SUF, we compare it with the baseline, maximum softmax probability (MSP) \cite{Hendrycks2016A}, w.r.t. ROC and PR in Fig. \ref{fige1}. DVN shares some nice properties of MSP, $e.g.,$ fixed classifier and single forward pass at the test stage. Moreover, DVN outperforms MSP by a large margin.

\textbf{Ablation studies}. To demonstrate the effectiveness of each module, we provide the detailed ablation study w.r.t. the choice of VAE/PixelCNN/Glow, disentanglement of $y$, modifying $p(z)$ to $q(z)$ with GAN and conditioned encoder.

$\bullet$ {PixelCNN/Glow-based DVN.} We also compared with the DVN that use pixel CNN or Glow rather than VAE as shown in Table 2. The pixelCNN/Glow-based DVN is consistently inferior than our solution. VAEs do have lower likelihood scores than Glow, but this gap is due to the different ways of computing likelihood of VAEs and flows. When computing the likelihood of a VAE, it is usually assumed that there is a unit Gaussian distribution at the output of the decoder. However the distribution of natural images is on a low dimensional manifold, so the likelihood number itself cannot be compared with Glow under this assumption. But VAEs are more robust than Glow due to the reason discussed in Sec 3.5, and in our experiments we found that Glows tend to put higher likelihood on OOD examples, which is bad for our usage.

$\bullet$ {Disentangling $y$ from $z$} is critical to our model. Table \ref{tableabcd} validates the contribution of this manipulation w.r.t. both threshold dependent and independent metrics. One can see that the DVN with disentanglement significantly outperforms its counterparts without disentanglement. This also implies the DVN has successfully learned to sufficiently minimize the mutual information between $z$ and $y$ to circumvent the challenge of conditioning $x$ on $y$.

\begin{table}[t]
\centering
\caption{The performance of DVN w/o disentanglement of $y$ from $z$ with ResNet backbone, and using CIFAR-10/SVHN as in-distribution/OOD, respectively.}
\resizebox{0.7\columnwidth}{!}{
\begin{tabular}{c|cc}
       \hline
Disentangle&TNR@TPR95\%&AUROC\\\hline\hline
$\surd$&\textbf{98.4}&\textbf{99.2}\\\hline
-&62.6&84.7\\
    \hline
    	\end{tabular}
}\label{tableabcd}
\end{table}

\begin{table}[t]
\centering
\caption{The performance of DVN w/o replace $p(z)$ with $q(z)$. We use ResNet backbone, and choose CIFAR-10/SVHN as in-distribution/OOD.}
\resizebox{0.6\columnwidth}{!}{
\begin{tabular}{c|cc}
    \hline
$q(z)$&TNR@TPR95\%&AUROC\\\hline\hline
$\surd$&\textbf{98.4}&\textbf{99.2}\\\hline
-&95.3&96.7\\       \hline
	\end{tabular}
} \label{tableab6} 
\end{table}

$\bullet$ {Without modifying $p(z)$ with $q(z)$}. Since modeling $p(x|y)$ is the core of DVN, we cannot remove $y$. Here, we give another ablation study that without modifying $p(z)$ with $q(z)$. As shown in Table \ref{tableab6}, there is a large margin between the DVN with or without disentanglement w.r.t. TNR@TPR95 and AUROC. The results demonstrate that disentangle $y$ from $z$ is of essential important for our framework.

$\bullet$ {Encoder condition on $y$}. In this paper we assume $q(z|x)=q(z|x,y)$, the motivation is during test time, $y$ may be a wrong label, we don't want it to confuse the encoder. Table \ref{tableency} gives a comparison of conditioning our encoder on $x$ or $(x,y)$.

\begin{table}[t]
\centering  
\caption{The performance of DVN use $q(z|x)$ and $q(z|x,y)$ encoder. We use ResNet backbone, and choose CIFAR-10/SVHN as in-distribution/OOD.}
\resizebox{0.7\columnwidth}{!}{
	\begin{tabular}{ccc}
	\hline
 &TNR@TPR95\%&AUROC\\\hline
$q(z|x)$&\textbf{98.4}&\textbf{99.2}\\\hline
$q(z|x,y)$&93.7&95.5\\\hline
	\end{tabular}
}\label{tableency}
\end{table}

\subsection{Detecting adversarial examples}

To detect adversarial examples, we train our DenseNet and ResNet-based classification network and DVN using the training set of CIFAR-10, CIFAR-100 or SVHN datasets, and their corresponding test sets are used as the positive samples for the test. Following the setting in \cite{Lee2018A}, we applied several attack methods to generate the negative samples, such as basic iterative method (BIM) \cite{kurakin2016adversarial}, Deepfool \cite{moosavi2016deepfool}, and Carlini-Wagner (CW) \cite{carlini2017adversarial}. The network structures are the same as for OOD verification.

We compare the DVN with the strategies in KD+PU \cite{feinman2017detecting}, LID \cite{ma2018characterizing}, SUF \cite{Lee2018A} in Tab. 4, and show that the DVN can achieve the state-of-the-art performance in most cases w.r.t. AUROC. In the "detection of unknown attack setting'', we can not access the adversarial examples of the test stage in the training or validation. Therefore, the previous works choose to use another attack generation method, $i.e.,$ fast gradient sign method (FGSM) \cite{goodfellow2014explaining}, to construct a validation set of adversarial examples. In here, we do not need another attack method as a reference, since the threshold of the DVN is only related to the validation set of in-distribution samples. Moreover, the pre-processing and model change as in \cite{Lee2018A} are not required in our proposed DVN. %The performance of dealing the adaptive attackers (be aware of deep verifier), spatially transformed adversarial attackers and unrestricted adversarial attackers are shown in the supplementary materials.

\begin{table}[t]
\centering
\caption{OOD verification results of image caption under different validation setups. We use CUB-200, LSUN and COCO as the OOD of Oxford-102, while using Oxford-102, LSUN and COCO as OOD of CUB-200.}
\resizebox{0.9\columnwidth}{!}{

\begin{tabular}{cc|ccc}
\hline
In-Dist&OOD&\multicolumn{3}{c}{Validation on OOD samples}\\\cline{3-5}
&& {TNR@TPR 95\%}  &{AUROC} & {Verif acc.}\\\hline\hline
&CUB&55.6&72.3&79.5\\
Oxford&LSUN&50.5&71.8&76.2\\
&COCO&40.3&74.4&73.3\\\hline\hline
&Oxford&39.8&68.4&72.5\\
CUB&LSUN&36.3&65.4&69.5\\
&COCO&35.4&60.7&71.0\\\hline
\end{tabular}
} 
\end{table}

\subsection{OOD for image captioning}

For detecting OOD samples in the image captioning task, we choose Oxford-102 and CUB-200 as the in-distribution datasets. Oxford-102 contains 8,189 images of 102 classes of flower. CUB-200 contains 200 bird species with 11,788 images. Each of them has 10 descriptions that are provided by \cite{reed2016learning}. For these two datasets, we use 80\%  of the samples to train our caption generator, and the remaining 20\% for testing in a cross-validation manner. The LSUN and Microsoft COCO datasets are used as our OOD dataset.

The captioner used in here is a classical image caption model \cite{xu2015show}. We choose the generator of GAN-INT-CLS \cite{reed2016generative} as our decoder's backbone, and replace its Normal distribution vector as the output of encoder $z$. A character level CNN-RNN model \cite{reed2016learning} is used for the text embedding which produces the 1,024-dimension vector given the description, and then projected to a 128-dimension code $c$. We configure the encoder and decoder with four convolutional layers and the latent vector $z$ is a 100-dimension vector. The input of the discriminator is the concatenation of $z$ and $c$, which results in a 228-dimension vector. A two-layer fully connected network with sigmoid output unit is used as the discriminator. Tab. {8} summarizes the performance of DVN in image caption task and can be regarded as a powerful baseline.

%% file: 5_Conclusion.tex
\section{Conclusions}

In this paper, we propose to enhance the performance of anomaly detection by verifying predictions of deep discriminative models using deep generative models. The idea is to train a conditional verifier network $q(x|y)$ as an approximation to the inverse posterior distribution. We propose Deep Verifier Networks (DVNs) which are based on a modified conditional variational auto-encoders with disentanglement constraints. We show our model is able to achieve state-of-the-art performance on benchmark OOD detection and adversarial example detection tasks.

\section{Acknowledgements}
This work was supported by the Jangsu Youth Programme [grant number BK20200238].